\documentclass[twoside,11pt]{article}

\usepackage[abbrvbib,accepted]{melba}
\usepackage{graphicx}

\usepackage{algorithm}
\usepackage{amsmath,amssymb}
\usepackage{cleveref}

\newtheorem{define}{Definition}

\usepackage{listings}
\floatstyle{ruled}
\newfloat{listing}{tb}{lst}{}
\floatname{listing}{Listing}
\usepackage{subcaption}

\melbaheading{2022:028}{https://www.melba-journal.org/papers/2022:028.html}{2022}{1-17}{06/2022}{11/2022}{Vlontzos, Rueckert and Kainz}{}{}

\ShortHeadings{A Review of Causality for Learning Algorithms in Medical Image Analysis}{Vlontzos, Rueckert and Kainz}
\firstpageno{1}

\title{A Review of Causality for Learning Algorithms in Medical Image Analysis}

\author{\name Athanasios Vlontzos\orcid{0000-0002-7672-2574} \email athanasios.vlontzos14@imperial.ac.uk \\  % start right after \author{, or there will be an extra space
	\addr Dept. of Computing, Imperial College London, London UK 
	\AND
	\name Daniel Rueckert\orcid{0000-0002-5683-5889} \email d.rueckert@imperial.ac.uk \\
	\addr TU M\"unchen, M\"unich, Germany \& Dept. of Computing, Imperial College London, London UK 
	\AND 
	\name Bernhard Kainz\orcid{0000-0002-7813-5023} \email b.kainz@imperial.ac.uk \\
	\addr FAU Erlangen-N\"urnberg, Erlangen, Germany \& Dept. of Computing, Imperial College London, London UK 
}

\begin{document}

% top matter
\maketitle
\abstract{
Medical image analysis is a vibrant research area that offers doctors and medical practitioners invaluable insight and the ability to accurately diagnose and monitor disease. Machine learning provides an additional boost for this area. However, machine learning for medical image analysis is particularly vulnerable to natural biases like domain shifts that affect algorithmic performance and robustness. In this paper we analyze machine learning for medical image analysis within the framework of Technology Readiness Levels and review how causal analysis methods can fill a gap when creating robust and adaptable medical image analysis algorithms. 
We review methods using causality in medical imaging AI/ML and find that causal analysis has the potential to mitigate critical problems for clinical translation but that uptake and clinical downstream research has been limited so far. 

%The recent developments in the field of Machine Learning have boosted the impact of automated  image analysis methods in real life medical practice. In this paper we are looking into how the inclusion of causality aides the development of robust, safe, fair and accurate algorithms. This is a short survey that looks into the current trends of the research field. 
}

\section{Introduction}

\begin{figure}
    \centering
    \includegraphics[width=\linewidth]{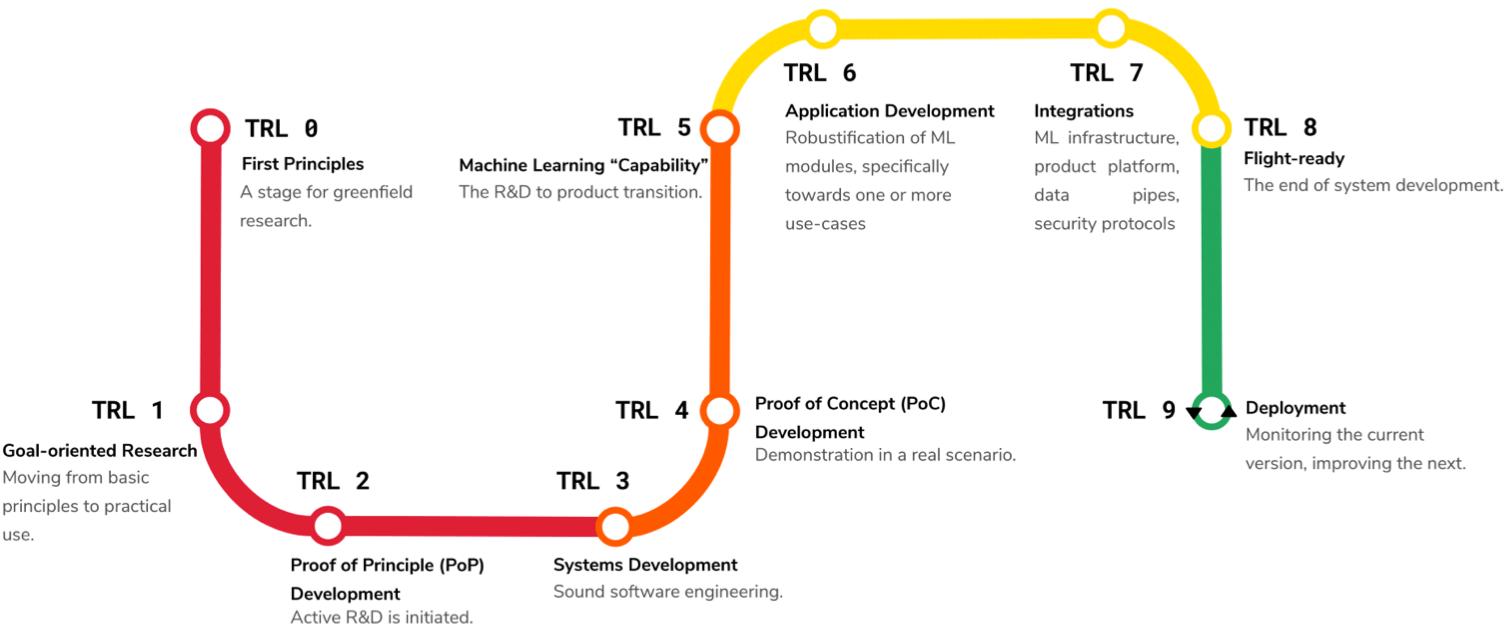}
    \caption{Technology Readiness Levels of ML Systems - most applications skip levels 5,6 where algorithms are made robust and production ready. Figure  reproduced from source without alterations; Source:~\citep{lavin2021technology} }
    \label{fig:mltrl}
\end{figure}

\begin{figure}
    \centering
      \includegraphics[width=\linewidth]{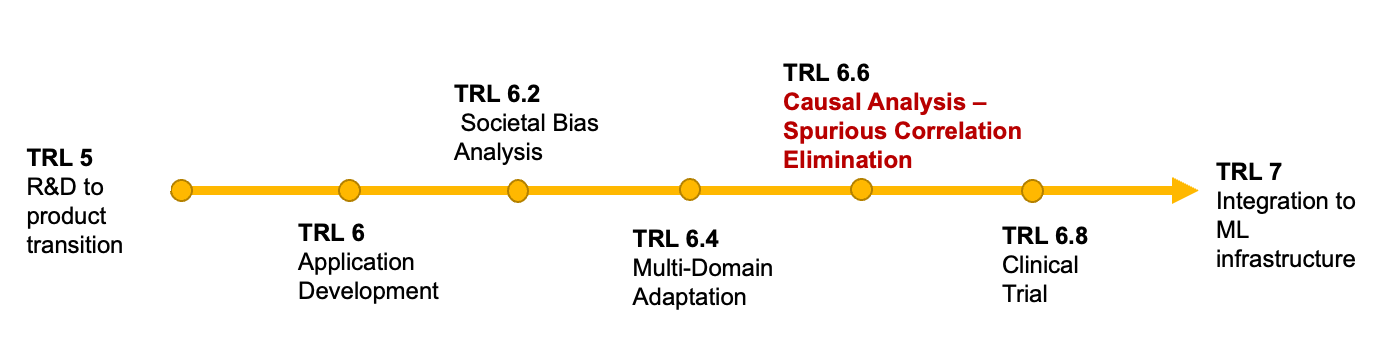}
          \caption{A in-depth view of the ML specific TRLs between TRL 6 and 7 in medical imaging - Inspired by~\citep{lavin2021technology}. In this paper we focus on the TRL 6.6 and argue the need and benefits of causally robust ML algorithms}
          \label{fig: our mltrl}
\end{figure}

%Recent advances in the field of computer science and more specifically Machine Learning (ML) have shown great potential in revolutionizing the applications relating to medical imaging. 
Medical imaging is an umbrella term encompassing a number of imaging techniques including Magnetic Resonance Imaging (MRI), X-ray imaging, Computed Tomography (CT), and Ultrasound (US) imaging , and is used primarily as supporting tool  for diagnosis and monitoring of diseases. From a computational perspective, the community has engaged in a wide variety of tasks concerning the automated interpretation of medical imaging and enabling a range of applications. Machine learning (ML) shows significant successes for applications like automated localization and delineation  of lesions and anatomies \citep{ronneberger2015u,kamnitsas2017efficient}  as well as for the automated  alignment of scans between patients and mapping of patient anatomy into canonical interpretation spaces \citep{maintz1998survey,haskins2020deep,grzech2020image,cabezas2011review}. 
Despite good in silico results, many of the approaches fail to translate into the clinical practice. While the reasons behind this can be complex and diverse, many have as a common factor the inability to adapt and be robust in clinical practice. Mapping this to the popular systems engineering framework of Technology Readiness Levels (TRL)~\citep{lavin2021technology} as shown in \Cref{fig:mltrl}, medical imaging AI/ML applications often skip from TRL 4 -- proof of concept--  to TRL 7 --deployment-- overlooking the very important TRLs 5 and 6 that make new systems robust to real world conditions. In \Cref{fig: our mltrl} we exhibit a finer grain view on the steps between TRL 6 and 7 that we deem to be important during the development of production-level medical imaging ML algorithms. 

Concerning is the inability of commonly used AI/ML medical imaging methods to differentiate between correlations and causation, making potentially deadly mistakes in the process. For example, \citep{DeGrave2020.09.13.20193565} identified a number of approaches that claim to have been able to diagnose COVID-19 from chest X-rays but ultimately fail to do so as they were instead picking up spurious correlations like hospital identifiers and the ethnicity of the patient. 

As pointed out by \citep{Castro2020,prosperi2020causal}, ML for medical imaging is susceptible to  different domain shifts that affect algorithmic performance and robustness in new environments. Population shifts occur when the train and test populations exhibit different characteristics that might include the prevalence of diseases, for example prevalence of lung nodules in polluted urban environments is higher than in a rural setting. An ML model trained in one of the two settings is not able to condition on the true causal links that are unseen in the images, hence, treats both populations the same way. %THis often leads to mis- or over- diagnosis of disease. 

Acquisition and annotation shifts effect the production of both images and ground truth labels used to train  ML algorithms. For example the same  MRI scanner used to acquire both images and annotations can lead to two different datasets, dependent on the scanner settings and medical beliefs of the radiologist performing the annotations. 

Finally, data selection biases are especially prevalent in the medical domains where expanded datasets can be hard to create for ethical, economic, and legal reasons. 

If acknowledged and mitigated by causal analysis, the aforementioned biases can help build robust and adaptable ML algorithms for medical imaging that minimize the chances of dangerous predictions due to spurious correlations. In essence, many of the negative phenomena seen in ML for medical imaging  could be solved if the community expands its involvement with the TRL points shown in \Cref{fig: our mltrl}, which suggest the inclusion of causal analysis as an  important step. However, causal analysis is not commonly used for the development of ML applications for medical imaging. 
Thus, in this review we explore recent research in this direction and the use of causal analysis for medical image machine learning applications. We looked at the major conferences - from example MICCAI, ISBI, IPMI from 2018 to April 2022- and journals of our field as well as some seminal preprints and included all that were directly related to both machine learning for medical imaging and causal reasoning. For a work to classify in our view, mentions of causality should transcend the discussion phase and be explicitly involved in the development or operation of the methods. %Recently, the first works that explicitly make use of causal analysis have appeared as peer-reviewed publications. 
In addition we will attempt to identify trends and lay out our beliefs for future directions of this field.

Throughout the following survey we assume that the reader is familiar with the notions of machine learning for medical imaging. While we invite the reader to refer to \citep{Pearl2009} for an in depth textbook of causality and \citep{yao2021survey,sanchez2022causal} for a survey in the general state of the art in causal ML and an opinion piece on the use of causality in medical machine learning respectively. We are attaching a small description and discussion of the key concepts related to this survey.

\section{Background}
We first introduce key concepts required for the upcoming discussion of causality-driven methods in medical image analysis. In \Cref{SCM} we are discussing the concepts of Structural Causal Models (SCM) and their parametrization as Directed Acyclical Graphs (DAG) as introduced  by J. Pearl \citep{Pearl2009}. In \Cref{PotentialOutcomes Theory} we are defining the notions behind Rubin's Potential Outcomes framework.

\subsection{Structural causal models \label{SCM}}

\begin{define}[Structural Causal Model] \label{functional causal model}
\label{scmdef}
A structural causal model (SCM) specifies a set of latent variables $U=\{u_1,\dots,u_m\}$ distributed as $P(U)$, a set of observable variables $=\{v_1,\dots, v_n\}$, a directed acyclic graph (DAG), called the \emph{causal structure} of the model, whose nodes are the variables $U\cup V$, a collection of functions $F=\{f_1,\dots, f_n\}$, such that $v_i = f_i(PA_i, u_i), \text{ for } i=1,\dots, n,$ where $PA$ denotes the parent observed nodes of an observed variable.
% in $G$ and $U_i$ is the set of node $v_i$'s latent parents.
% The latent noise term appearing in each $f_i$ can be suppressed into $PA_i$ by enforcing the convention that every observed node has an independent latent variable as a parent in $G$. This convention is adopted throughout the following.
\end{define}

%Definition~\ref{scmdef} deviates slightly from the conventional definition, and allows $f_i$ to take in a set of latent nodes as a parent. This means that an observed node can have multiple latent parents, and a latent node can have multiple observed children. %A specific formulation---the Markovian SCM, or \emph{Functional Causal Model}---has a single latent parent per observed node, and stipulates that latent variables be mutually independent. This is the case on which we focus in this paper.

The collection of functions $F$ and the distribution over latent variables induces a distribution over observable variables: $P(V=v) := \sum_{\{u_i \mid f_i(PA_i, u_i)\,=\,v_i\}}\\ P(u_i)$ %\,\forall\, u_i \in\, U.$ 
In this manner, we can assign uncertainty over observable variables despite the fact the underlying dynamics are deterministic. 
%An example causal structure, represented as a directed acyclic graph (DAG), is depicted in \ref{examplescm}.

% \begin{define}[Submodel]
% Let $M$ be a structural causal model, $X$ a subset of observed variables with realization $x$. A submodel $M_x$ is the causal model with the same latent and observed variables as $M$, but with functions replaced with
% $F_{x} = \{f_i \mid v_i\notin X\} \cup \{f_{j}^{'}(PA_j, u_j) := x_j\ |\ v_j \in X\}$.
% \end{define}

% \begin{define}[do-operator]
% Let $M$ be a structural causal model, $X$ a set of observed variables. The effect of action $\text{do}(X=x)$ on $M$ is given by the submodel $M_{x}$.
% \end{define}

Moreover, the $do$-operator forces variables to take certain values, regardless of the original causal mechanism. Graphically, $do(X=x)$ means deleting edges incoming to $X$ and setting $X=x$. Probabilities involving $do(x)$ are normal probabilities in submodel $M_x$: $P(Y=y \mid \text{do}(X=x)) = P_{M_x} (y)$.

\noindent\textbf{Counterfactual inference} \label{Section: two approaches to counterfactual inference}
The latent distribution $P(U)$ allows one to define probabilities of counterfactual queries, $P(Y_{y}=y) = \sum_{u \mid Y_{x}(u)=y} P(u).$ For $x \neq x'$ one can also define joint counterfactual probabilities,
$P(Y_{x}=y, Y_{x'}=y') = \sum_{u \mid Y_{x}(u)=y,\text{ }\& Y_{x'}(u)=y'} P(u).$
Moreover, one can define a counterfactual distribution given seemingly contradictory evidence. Given a set of observed evidence variables $E$, consider the probability $ P(Y_{{x}}={y}' \mid E = {e})$.

\begin{define}[Counterfactual]\label{def:counterfactual}
% \noindent\textbf{Definition 2(Counterfactual):}\emph{
The counterfactual sentence ``$Y$ would be $y$ (in situation $U=u$), had $X$ been $x$'', denoted $Y_{x}(u) = y$, corresponds to $Y=y$ in submodel $M_x$ for $U = u$.
% }
\end{define}

\begin{figure}[t] 
\centering

\subfloat[
    \label{examplescm}]{
        \includegraphics[width=0.17\linewidth]{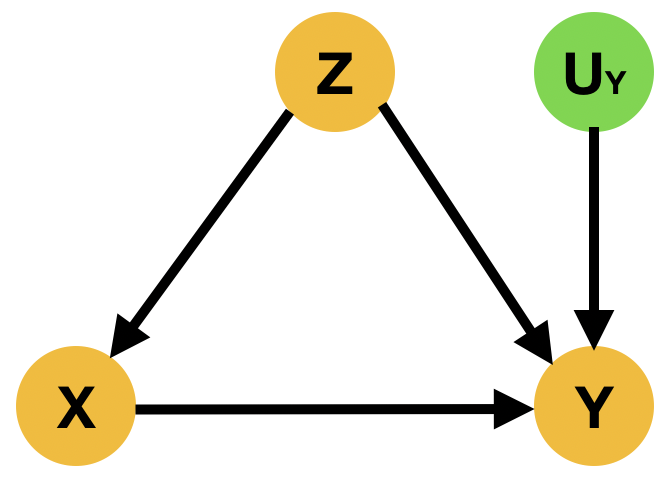}
        }
\subfloat[
    \label{examplescm_twin}]{
        \includegraphics[width=0.17\linewidth]{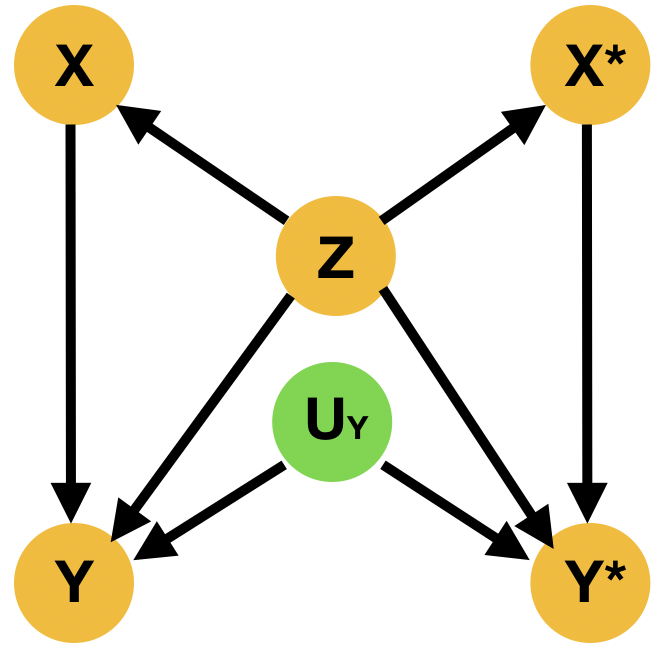}
        }
\subfloat[
    \label{examplescm_twin_do_x}]{
        \includegraphics[width=0.17\linewidth]{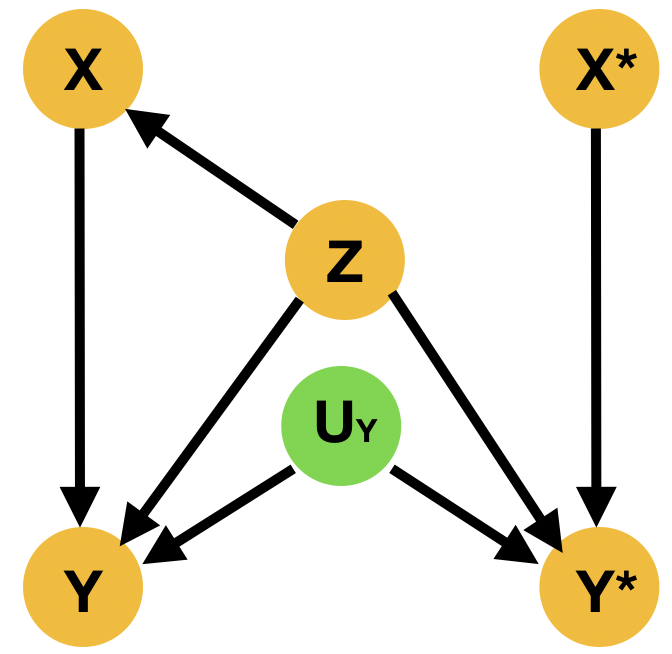}
        }
\subfloat[ 
    \label{examplescm_twin_do_x_2}]{
        \includegraphics[width=0.17\linewidth]{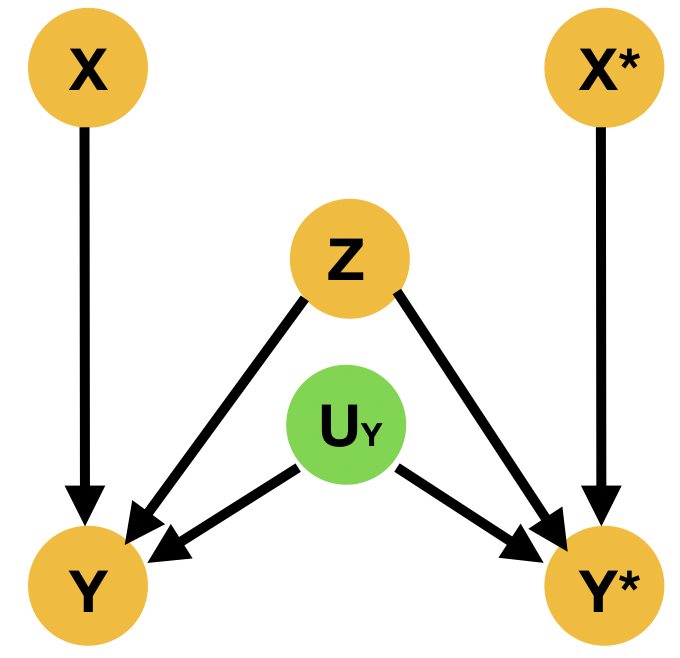}
        }
\subfloat[
    \label{figure: toy example}]{
        \includegraphics[width=0.17\linewidth]{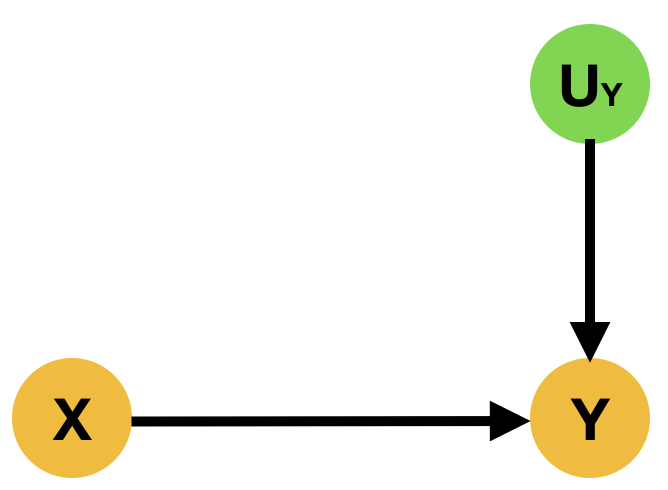}
        }
\caption{Orange nodes are observed, green latent. (a) Example SCM; (b) twin network of (a); (c) intervention in the twin network on node $X^*$; (d) interventions in the twin network on $X$ \& $X^*$; (e) Uncounfounded version of (a).
}
\label{figure: twin networks}
\end{figure}

%, which reads ``what is the probability that $Y$ would have been ${y}'$, given that $E$ was observed to be ${e}$, if $X$ had been ${x}$?''
% In the case of binary variables, this query is related to the probability of necessity and sufficiency \citep{pearl1999probabilities} \todo{I am sorry, not sure why we mention it here, unless we want to elaborate on this?}.
Despite the fact that this query may involve interventions that contradict the evidence, it is well-defined, as the intervention specifies a new submodel. Indeed, $ P(Y_{x}={y}' \mid E = {e})$ is given by~\citep{Pearl2009}
$ \sum_{{u}}
P(Y_{{x}}({u})={y}')P({u}|{e})\,.$
There are two main ways to resolve this type of questions; the Abduction-Action-Prediction paradigm and the Twin Network paradigm shown  respectively in ML literature among others in \citep{castro2020causality,vlontzos2021estimating}. In short given SCM $M$ with latent distribution $P(U)$ and evidence ${e}$, the conditional probability $P(Y_{{x}} \mid {e})$ is evaluated as follows: 1) \emph{Abduction:} Infer the posterior of the latent variables with evidence ${e}$ to obtain $P(U \mid {e})$, 2) \emph{Action:} Apply $\text{do}({x})$ to obtain submodel $M_{x}$, 3) \emph{Prediction:} Compute the probability of $Y$ in the submodel $M_{x}$ with $P(U \mid {e})$. Meanwhile the Twin Network paradigm casts the resolution of counterfactual queries to Bayesian feed-forward inference by extending the SCM to represent both factual and counterfactual worlds at once~\citep{balke1994counterfactual}. We exhibit an illustration of the twin network paradigm in \Cref{figure: twin networks}

\subsection{Potential outcomes -- Average treatment effect\label{PotentialOutcomes Theory}}
The potential outcomes framework introduced by \citep{10.1214/ss/1177012031} and \citep{10.1214/aos/1176344064} explore the potential outcomes of a given intervention. Formally the framework is primarily interested in $Y(a), a \in \mathcal{C}$ where $Y$ is the outcome given intervention $a$ which belongs in the set of possible actions $\mathcal{C}$. For the simple case where we are interested in the outcome of a specific treatment, we can use the following definitions:

\begin{define}[Treatment under Potential Outcomes]
$\mathcal{D}_i$: Indicator of treatment intake for unit $i$
\begin{equation}
\mathcal{D}_i= \begin{cases}
 1 , & \text{if unit $i$ received the treatment } \\
  0, & \text{otherwise}\end{cases}
\end{equation}

\end{define}

\begin{define}[Potential Outcomes]
$\mathcal{Y}_{di}$: Potential outcomes for unit $i$ depending if treatment has been applied or not

\begin{equation}
 \mathcal{Y}_{di}= \begin{cases}
 \mathcal{Y}_{1,i}, & \text{Potential Outcome for unit $i$ receiving the treatment } \\
  \mathcal{Y}_{0,i}, & \text{Potential Outcome for unit $i$ not receiving the treatment }\end{cases}
\end{equation}

\end{define}

Using these quantities we are able to define the causal effect 

\begin{define}[Causal Effect]

Causal Effect of the treatment on the outcome for unit $i$ is the difference between its two potential outcomes:
\begin{equation} \label{treatment_effect_def}
    \tau_i = \mathcal{Y}_{1,i} -  \mathcal{Y}_{0,i}
\end{equation}
\end{define}

In \Cref{treatment_effect_def} we define the causal treatment effect for an individual unit $i$. We can also define the average treatment effect that looks into a group of individual units $i \in \mathcal{G}$.

\begin{define}[Average Treatment Effect]

$\tau_{ATE}$  is the difference between all treatment potential outcomes and all control potential outcomes

\begin{equation} \label{ATE_def}
\begin{split}
      \tau_{ATE}    & = \frac{1}{N} \sum^{N}_{i}\mathcal{Y}_{1,i} - \frac{1}{N} \sum^{N}_{i}\mathcal{Y}_{0,i} \\
         & = E[\mathcal{Y}_{1,i}-\mathcal{Y}_{0,i}]\\
         & = E[\tau_i]
   \end{split}
\end{equation}
\end{define}

In literature we can find many variation of this measure like the individual treatment effect (eg. \citep{arxiv.2104.13730}) where we look on the treatment effect on an individual rather than aggregate over a population. In the interest of conciseness we will not be be exploring the full range of possible variations upon the ATE in this survey but just acknowledge that they exist and call upon them depending on the method we are discussing. 

One last concept required for our discussion is \emph{propensity score}. Introduced by \citep{rosenbaum1983central} it is defined to be the probability of treatment assignment conditioned on observed covariates. Mathematically it can be described as $e = P(T \mid X)$ where $X$ are the covariates and $T$ the treatment. Commonly used as a matching criterion to form sets of treated and untreated subjects that are close in the covariate space; these sets are subsequently used to estimate the causal effect of the treatment.

\section{Causal discovery in medical imaging}
Causal discovery is an open and challenging research problem, directly touching the most fundamental aspects of scientific exploration; the discovery of causal relations~\citep{arxiv210302582}. In this setting we are trying to estimate the mechanisms that describe the causal links between variables from data. In order to make the problem of causal discovery tractable, common causal discovery methods make a series of assumptions that can be summarized as: 
\begin{itemize}
    \item Acyclicity: we are able to describe the causal structure as a Directed Acyclical Graph (DAG)
    \item Markovian: all nodes are independent of their non-descendant when conditioned on their parents
    \item Faithfulness: all conditional independences are represented in the DAG
    \item Sufficiency: any pair of nodes in the DAG have no common external causes
\end{itemize}
Moreover, the vast majority of approaches tackling causal discovery formulate the problem as graph modeling challenge. We refer the reader to \citep{10.3389/fgene.2019.00524,arxiv210302582} for a more thorough review of causal discovery methods based on graphical models. For the purposes of this survey we note that  methods are often categorized based on their approach into \emph{constraint-based}, \emph{score-based} and \emph{optimization-based}~\cite{zheng2018dags} methods.

Constraint-based methods relate to approaches like \citep{spirtes2000causation} PC and Fast Causal Inference (FCI) that test conditional independence between factors to assess their causal links. The main intuition is that two statistically independent variables are not causally linked. First, pairwise independence is evaluate to determine the undirected structure. Following this conditional independence is tested to orient the links between the nodes. If two nodes fail this test, then they can be added to the separation set of each other used to orient colliders - nodes of the causal graph with two or more incoming links. The main difference between the aforementioned PC and FCI algorithms is FCI's ability to be asymptotically correct in the presence of confounders. Both of them are however limited to the causal \emph{equivalence classes}, \emph{i.e.}, causal structures that satisfy the same conditional Independence. A method that searches over the space of possible equivalence classes is the Greedy Equivalence Search (GES)~\citep{10.1162/153244303321897717} officially considered a score-based method which uses the Bayesian Information Criterion (BIC). Similarly, \cite{pamfil2020dynotears} use a score based approach that characterizes the acyclicity constraint of Directed Acyclical Graphs as a smooth equality constraint.

In the computer vision field, visual causal discovery has been spearheaded by tasks related to the CLEVRER dataset~\citep{yi2020clevrer} where ML algorithms are asked to understand a video scene and answer counterfactual questions.\citep{li2020causal} learn to predict causal links from videos, by parameterizing the causal links as strings and springs where the algorithm predicts their parameters. \citep{lowe2022amortized} perform a similar task by inferring stationary causal graphs from videos. Furthermore, \citep{nauta2019causal} develop a causal discovery method based upon the use of attention-enabled convolutional neural networks. \citep{2204.04875} address the task of causal discovery by training a neural network to induce causal dependencies by predicting causal adjacency matrices between variables. Finally, works like \citep{gerstenberg2021counterfactual} analyze how humans perform the task of causal discovery in relation to how machine learning approaches the same task. 

In the sub-field of medical imaging, causal discovery has not been explored to its fullest potential. Most works involve functional Magnetic Resonance Imaging (f-MRI). f-MRI is able to highlight active areas of the brain, and as during different physiological processes (working or resting) activate a series of brain areas. Causal discovery in f-MRI attempts to identify causal links among neural processes. \citep{Sanchez-Romero245936,sanchez2019identification} diverge from the usual DAG paradigm and introduce the Fast Adjacency Skewness (FASK) method which exploits non-Gaussian features of blood-oxygen-level-dependent (BOLD) signals to identify causal feedback mechanisms. 
In a very interesting work,  \citep{HuangScholkopf2020} parametrize both causal discovery in f-MRI and domain adaptation as a non-stationary causal task that they then proceed to solve. In the field of medicine but not imaging \citep{mani2000causal} extract causal relations from medical textual data.

\citep{bielczyk2017causal} looked into applications on f-MRI and extracting causal relations between neural processes by applying the most common causal inference techniques including dynamic causal modeling, Bayesian networks, transfer entropy and ranked them based on their perceived suitability and performance on the tasks at hand.  Similarly, \citep{cortes2021going} explore deep learning methods for schizophrenia analysis and outcome prediction, mostly with the use of f-MRI inputs, and argue for the need of causally enabled ML methods to  produce plausible hypothesis explaining observed phenomena. Recently \citep{10.1007/978-3-031-16431-6_26} propose deep stacking networks (DSNs), with adaptive convolutional kernels (ACKs) as component parts to aid the identification of non-linear causal relationships.

Furthermore, \citep{10.3389/fgene.2018.00347} discuss bivariate causal discovery for imaging data. The authors develop a non-linear additive noise model for that they show a causal discovery task with genetic data and for causal inference in the Alzheimer's Disease Neuroimaging Initiative (ADNI) MRI data.  \citep{reinhold2021structural} propose a Structural causal model that encodes causal functional relationships between demographics and disease covariates with MRI images of the brain in an attempt to identify Multiple Sclerosis (MS). 

While the field of medicine is governed by causal relations between physiological processes, little work has been done in the field of medical imaging and causal discovery. We believe that this is mostly due to the significant inherent difficulty of the task at hand, in conjunction with the lack of the required meta-data that are needed to characterize causal links that are not visible in the image. In other words, we do not think that images by themselves possess enough information for identification of causal links but we are adamant that they can serve as a useful tool and source of extra information when combined with medical knowledge. It is, however, important to note the potential the aforementioned time-series based causal discovery methods exhibit as f-MRIs can themselves be thought of as time-modulated events, and be used in conjunction with other time-based modalities.

Finally, relating back to \Cref{fig: our mltrl}, causal discovery can aid bring to light causal links that were not previously known. It can further help, probe the beliefs of other algorithms and hence root out spurious correlations or the encoding of societal biases that their creators carried. As such we believe that causal discovery can assist the completion of TRLs 6.6 through audits of the practitioners beliefs. Most importantly, however, in the subsequent sections we will be discussing causal inference, that assumes the existence of high quality causal diagrams. Causal discovery is responsible for establishing these diagrams and hence indirectly contributing  significantly at TRLs 6-6.4. 

\section{Causal Inference in medical imaging}

While causal discovery using medical images is limited to some applications involving f-MRI scans, causal inference is significantly more active as an area of research. Highlighting its importance, \citep{Castro2020} argue that causal inference can be used to alleviate some of the most prominent problems in medical imaging. They argue that acquisition and annotation of medical images can exhibit bias from the annotators and curators of the datasets. As such, causality aware methods can learn to account for such biases and reduce their effects. Moreover, as the training datasets represent a limited population with specific characteristics, medical imaging algorithms are susceptible to population, selection and prevalence biases if not properly controlled for these variations. These biases could for example arise when an algorithm is trained by a vast majority on data that come from a given geographical region $X$,  then it implicitly learns the prevalence of diseases for that group; if it is deployed in a different region with a population that is characterized by different genotype and phenotype characteristics, the biases that the algorithm has learned could lead to mis- or under- diagnosis of diseases. 

In our exploration of the use of causal inference in medical imaging literature we identified five main sub-fields of research that leverage causal insights. We found that causal inference is overwhelmingly used  to contribute to fairness, safety and explainability of the existing approaches. There are some albeit limited uses in generative modeling of medical images, domain generalization and out-of-distribution detection. As we will expand in the following sections, we believe these areas are ready for more applications of causal inference. 

We note that all the works mentioned below have causality as a key part of their proposed methodologies. We extended our survey not only to peer reviewed publications but also to notable preprints that appear to have produced significant discussion in the machine learning medical imaging community. 

\subsection{ Medical Analysis}
We continue our survey with causally enabled methods for medical analysis that utilize imaging data.
% how causal inference can help medical imaging algorithms perform general medical analysis. 

 In \citep{10.1007/978-3-030-87199-4_17} the authors use a normalizing flow-based causal model similar to \citep{pawlowski2020deep} in order to harmonize heterogeneous medical data. Applied to T1 brain MRI for the classification of Alzheimer's disease, the method abides by the abduction-action-prediction paradigm to infer counterfactuals which are then used to harmonize the medical data.
\citep{10.1007/978-3-030-78191-0_4} circumvent the identifiability condition that all confounders have to be known and measured by leveraging the dependencies between causes in order to determine  substitute confounders; they apply their method in brain neuroimaging for Alzheimer's disease detection.
On a similar note, 
\citep{10.1007/978-3-030-78191-0_5} issue an alternative to expectation maximization(EM) for dynamic causal modeling in f-MRI brain scans. Their approach is based upon the multiple-shooting method to estimate the parameters of ordinary differential equations (ODEs) under noisy observations required for brain causal modeling. The authors suggest an augmentation of the aforementioned method called multiple-shooting adjoint method by using the adjoint method to calculate the loss and gradients of their model. 

 \citep{clivio2022neural} propose a neural score matching method for high dimensional data that could be potentially very helpful for the development of medical imaging applications as they develop methods for causal inference in high dimensional settings; allowing thus the use of medical images in a more straightforward way that avoids pre-processing them to a lower dimensional latent space. Finally, \citep{ramsey2010six} identified six problems that causal inference can assist in solving in the field of functional MRI analysis.

\subsection{Fairness, Safety \& Explainability}
Another application of causal inference in the field of medical imaging revolves around fairness, safety and explainability, directly related to TRLs 6.2 6.6 and 6.8. Medical tools like medical imaging analysis have a significant impact on the well-being of people affected by their use. Doctors and patients alike need to be able to trust the AI/ML methods in order to use them, while contrary to other AI/ML applications unwanted bias and poor performance can often be deadly. As such, the need to have fair, safe and explainable algorithms arises. Causal inference is a great tool to analyze black box AI/ML methods and make sure that they are not carrying unwanted societal biases and mitigate any robustness problems that might arise. 

In this field \citep{kayser2020understanding} accompany their ML algorithm to detect polyps in human intestines with a causality inspired analysis on the effectiveness of their method. Similarly \citep{da2020biomechanical} use generative model produced brain MRI images of brain atrophy to evaluate and explore different causal hypothesis on brain growth and atrophy. On a similar note  \citep{li2021searching} develop a causal inference-based method to search associations in genomics data from the UK Biobank.
Additionally,  \citep{baniasadi2021dbsegment} employ causal analysis to explain the performance of their brain structure segmentation network.
 Similarly  \citep{10.1007/978-3-030-87199-4_49}  employ mediation analysis to identify the units and parameters of radiological reports that influence their classifier's outcomes; this method is applied on chest X-rays. In their work   \citep{zapaishchykova2021interpretable} develop a Bayesian causal model to interpret the outputs and functionality of their Faster-RCNN based pelvic fracture classifier in CT images. Their Bayesian causal model matches lower confidence predictions with higher confidence ones and then updates the prediction set based on these matching.   \citep{garciaeffect} explore the effect of uncontrolled confounders in medical imaging applications and observe that regardless of task and architecture, total confounding can be used to explain the difference in performance between development of the models and real life applications.  \citep{adebayo2022post} evaluate new metrics to quantify the effect of spurious correlations in age regression from hand X-rays. They show that only under certain conditions these metrics can be trusted and call for a paradigm shift in the effort to identify spurious correlations

Concerning fairness, \citep{chen2021algorithm} discuss the effects of biases in medical ML and how biases like image acquisition, genetic variation, and intra-observer labeling result in healthcare disparities. They go on to argue that causal analysis in medical ML can greatly help mitigate such biases. Expanding on this argumentation, \citep{holzinger2022information} argue that information fusion is key to achieve greater transparency and safety in medical imaging ML applications. In a slightly different position paper,  \citep{santa2021need} argue for the standardization of medical metadata in order to assist causal inference techniques in biomedical ML. Along similar lines,   \citep{garcea2021use} argues for the use of causal intuition when designing medical imaging datasets. Meanwhile,  \citep{causalprobe} uses causal inference to estimate the necessity and sufficiency of the type and quantity of data to include in a medical imaging dataset in order to improve model performance under strict computational and financial constraints. In addition, \citep{vlontzos2021next} contend that causal analysis can help alleviate biases and provide the necessary trust to medical imaging application in deep space manned missions. 

Finally, \citep{bernhardt2022investigating}  investigated questions of algorithmic fairness in medical imaging ML under a causal prism, focusing on the issue of under-diagnosis they highlighted some issues that warrant more attention in prior pieces of literature. \citep{schrouff2022maintaining} perform a thorough evaluation of the biases and unfairness that can arise in cross-hospital deployment of medical ML solutions asserting a causal analysis as a potential method to alleviate these issues. \citep{benkarim2021cost} use propensity scores (see \cref{PotentialOutcomes Theory}) to quantify diversity due to major sources of population stratification and hence assess fairness.

\subsection{Generative methods }
Generative modeling attempts to learn variable interdependencies such that the model is able to generate realistic samples that abide by certain characteristics aiding the admittance past TRLs 6.4 and 6.8. Variational Autoencoders \citep{kingma2013auto}, Generative adversarial networks~\citep{goodfellow2014generative} and normalizing flows~\citep{dinh2016density} are examples for approaches that try to estimate the underlying data distribution from which they then sample to produce new data. Causal inference in generative modeling is a relatively underdeveloped field, especially in the context of medical imaging due to the inherit difficulty to acquire good quality training signals for the counterfactual samples.

 \citep{gordaliza2022translational} develop a two stage methodology where Tuberculosis infected lung CT images are analyzed in a disentangled manner and produce counterfactual images depicting how the patient would look like if they were healthy. Contrary to other approaches the authors use a DAG to represent the image generation process and parametrize it using a neural network such that sampling and use of it is straightforward for the counterfactual generation  step. 
\citep{pawlowski2020deep} developed a normalizing flow model to perform the abduction step in an abduction-action-prediction counterfactual inference task and are able to generate plausible brain MRI volumes. Reynaud et al. in ~ \citep{dartagnan} assume a different approach and develop a generative model based on Deep Twin Networks~\citep{vlontzos2021estimating}. Performing counterfactual inference in the latent space embeddings, the authors are able to generate realistic Ultrasound Videos with different Left Ventricle Ejection Fractions. Their approach is similar to \citep{kocaoglu2018causalgan} in the sense that a GAN is used to provide a training signal for the generated, counterfactual samples. 
 Moreover, \citep{kumar2022counterfactual} in a methodologically similar note, generate counterfactual images to guide the discovery of medical biomarkers in brain MRI volumes. 
 
 Finally \citep{sanchez2022healthy} use deep diffusion model to ask counterfactual questions and generate hard to obtain medical scans. These, in turn, are used to augment existing datasets for other downstream tasks.

\subsection{Domain Generalization}
One of the most promising areas where causal reasoning can be applied in the field of medical imaging is Domain Adaptation and Out-of-Distribution detection, directly associated with TRL 6.4. If we model the generative process that results in a medical image and include factors like the medical history, the disease, imaging domain, etc. we can then go on and interpret domain generalization and adaptation as a model that is able to perform well under different treatments in the imaging domain parameter, as argued by \citep{HuangScholkopf2020}. In their paper Huang et al model domain adaptation as a non stationary change in the underlying causal graph and propose methods to identify and resolve these changes. 
\citep{10.1145/3450439.3451878} analyze the domain shifts experienced in clinical deployment of AIML algorithms from a causal perspective and then proceed to investigate and benchmark eight popular methods of domain generalization. They find that domain generalization methods fail to provide any improvement in performance over empirical risk minimization in situations where we find sampling bias. Similarly \citep{fehr2022causal} model the causal relationships leading to the medical images and create synthetic datasets in order to evaluate the transportability of methods to external settings where interventions on factors like ages, sex and medical metrics have been performed.

\citep{ouyang2021causality} apply a causal analysis on the problem of domain generalization in segmentation of medical images. They first simulate shifted domain images via a randomly weighted shallow network; then they intervene upon the images such that spurious correlations are removed and finally train their segmentation model while enforcing a domain invariance condition. \citep{valvano2021re} develop a method to reuse adversarial mask discriminators for test-time training to combat distribution shifts in medical image segmentation tasks. In their discussion of their method they explain the good performance of their method under a causal lens. Finally \citep{10.1007/978-3-030-20351-1_27} build a causal Bayesian prior to aide MRI tissue segmentation to generalize across different medical centers.

\subsection{Out of Distribution Robustness \& Detection}
We ought to consider the use of causal reasoning and inference on out of distribution detection tasks aiding TRLs 6.6 and 6.8. Commonly seen as anomaly detection tasks many methodologies attempt to learn the underlying distribution of ``normal"- \emph{in-distribution} data and assess whether test \emph{out-of-distribution} ``pathological" samples belong to the same distribution or not. In a related task, researchers sometimes treat the \emph{out-of-distribution} as a robustness criterion and attempt to develop methodologies that can operate equally well in all domains.
It is evident that an alternative, where we learn the causal dependencies that make samples that are considered \emph{out-of-distribution} can yield significant benefits in the field of anomaly detection. In this sub-field we only found a few works exploring this approach in medical imaging but we believe that more works will appear once the community gets more familiar with the benefits of causal reasoning. 

\citep{ye2021ood} give a causal explanation to diversity and correlation shifts and proceed to benchmark out-of-distribution methods, showing that the aforementioned shifts are the main components of the distribution shifts found in OOD datasets. Similarly  \citep{ye2021out}, introduce an influence function and a novel metric to evaluate OOD while analyzing their contributions from a causal standpoint. 
\citep{liu2021learning} propose a causal semantic generative model in order to address OOD prediction from a single training domain. Utilizing the causal invariance principle they disentangle the semantic causes of prediction and other variation factors achieving impressive results. 

On the robustness of medical procedures \citep{10.1007/978-3-031-16449-1_37}, built a causal tool segmentation  model that iteratively aligns  tool masks with observations.  Unable to deal with occlusions and without leveraging temporal information the authors of this recent work also comment on the future next steps of robust causal machine learning tools. 

Even though not directly applied in medical imaging, causality has been playing an important role in the algorithmic robustness literature as seen in \citep{Zhang2022AdversarialRT,papangelou2018toward}. Meanwhile the machine learning for medical imaging community has shown great interest in developing robust algorithms, \citep{hirano2021universal,10.1007/978-3-658-25326-4_6}. We hope that these two communities will soon come together and use causal reasoning in making machine learning for medical imaging algorithms robust.

\section{Discussion and Conclusion }
We have identified a wide range of possible applications of causality in medical imaging. However, we have not yet observed an enthusiastic uptake of causal reasoning and causal considerations from the community. We believe that introducing causal reasoning in medical imaging applications would benefit both the performance and robustness of the algorithms but most of all would provide the required scrutiny and security that doctors demand from their tools. A model that is able to be probed causally and does not operate as a black box can be trusted more easily by healthcare professionals since they will be able to understand its inner workings; enabling safe human-machine decision making  \citep{BUDD2021102062}. Moreover it would ameliorate the legal hurdle of accountability  that developers of medical ML applications face, as we would be able to  explain the processes and logic of the models. As machine learning algorithms in medical imaging make their way towards clinical practice we hope to see more causal machine learning solutions being trialed. 

Furthermore, as medical and medical imaging applications move away from  large hospitals, towards first responders, remote community doctors and even astronauts,  we need tools that are robust to extreme circumstance differences. Causally enabled ML has shown great promise in its ability to adapt as it does not depend on correlations that might or might not exist in the new domain, but, like humans, it taps into the underlying causal relationships that are very hard to shift. That being said, causal ML is not a panacea; our theoretical capabilities restrict absolute certainty to a limited number of well defined conditions that Causal Diagrams have to obey. When shifting focus to real life scenarios, we are forced to accept trade offs and be considerate about the limitations of our approaches. 

Causally enabled ML for medical imaging is a very promising research field that the authors of this survey believe is vital for  next generation of tools for medical imaging. We are very excited about what the future of this research might bring and we hope to have inspired researchers and practitioners with this review  to consider the causal ML route for medical imaging research and application development.

\bibliography{bibliography.bib}

\end{document}